\pdfoutput=1

\documentclass[11pt]{article}

\usepackage{acl}

\usepackage{times}
\usepackage{graphicx}
\usepackage{latexsym}
\usepackage{tablefootnote}
\usepackage{hyperref}

\usepackage[T1]{fontenc}

\usepackage[utf8]{inputenc}

\usepackage{microtype}

\usepackage{inconsolata}

\title{The Naughtyformer: A Transformer Understands Offensive Humor}

\author{Leonard Tang \hspace{0.3cm} Alexander Cai \hspace{0.3cm} Steve Li \hspace{0.3cm} Jason Wang \\
        Harvard University \\ Cambridge, MA 
        }

\begin{document}
\maketitle
\begin{abstract}
Jokes are intentionally written to be funny, but not all jokes are created the same. Some jokes may be fit for a classroom of kindergarteners, but others are best reserved for a more mature audience. While recent work has shown impressive results on humor detection in text, here we instead investigate the more nuanced task of detecting humor \textit{subtypes}, especially of the less innocent variety. To that end, we introduce a novel jokes dataset filtered from Reddit and solve the subtype classification task using a finetuned Transformer dubbed the Naughtyformer. Moreover, we show that our model is significantly better at detecting offensiveness in jokes compared to state-of-the-art methods. 

\end{abstract}


\section{Introduction}

The field of humor detection has received much interest over the years. Early work attempted to leverage N-grams \cite{Taylor04computationallyrecognizing}, stylistic features of humor (\citealp{Mihalcea2006LEARNINGTL}), and Random Forest classifiers acting on Word2Vec embeddings and Human Centric Features \cite{Yang2015HumorRA} to detect the presence of humor in text. More recently, deep learning-based approaches have been explored for humor detection, including Convolutional Neural Networks combined with Highway Networks \cite{chen-soo-2018-humor}, Long Short Term Memory and Gated Recurrent Unit Networks \cite{deOliveira2015HumorDI}. With the popularization of the Transformer architecture \cite{Vaswani2017AttentionIA}, researchers have turned towards finetuning large language models on the downstream task of humor classification (\citealp{Blinov2019LargeDA}; \citealp{Weller2019HumorDA}; \citealp{Annamoradnejad2020ColBERTUB}).


Continuing in this vein, \citet{Peyrard2021LaughingHC} leverage large language models to discriminate serious and humorous sentences from a challenging dataset while discovering evidence for humor-sensitive Transformer attention heads. Contemporaneously, \citet{Pitsilis2018DetectingOL} demonstrate that Recurrent Neural Networks are capable of detecting offensive language in Tweets drawn from Twitter. \citet{barbieri2020tweeteval} improve upon these results by leveraging Transformer-based architectures. 

Despite this progress, no prior work exists on the more nuanced task of classification amongst humor subtypes. We contend that this is a worthwhile domain to explore, especially when considering humor subtypes of the more offensive variety. In particular, the delineation between humor and offensive speech is often a blurry one, and ideally it should be possible to discern if a joke has been taken too far to prevent offending others. In order to do so, we curate and introduce a comprehensive dataset of offensive jokes and perform extensive modelling experiments using Transformers. 


\begin{table}[htbp]
    \centering
    \resizebox{\columnwidth}{!}{
    {\renewcommand{\arraystretch}{1.2}
    \begin{tabular}{lllll}
        \hline
        \textbf{Statistic} & \textbf{Clean} & \textbf{Dark} & \textbf{Dirty} & \textbf{News} \\
        \hline
         Examples & 7450 & 79230 & 5473 & 10710 \\
        Avg. Length & 31.47 & 24.64 & 55.24 & 778.84 \\
        SD. of Length & 46.21 & 78.67 & 91.14 & 292.34\\
        Avg. Upvotes & 87.30 & 105.11 & 38.85 & N/A\tablefootnote[1]{Note that Thomson Reuters news articles do not contain any notion of upvotes, as they are not embedded in a social media platform like Reddit.}\\
        SD. of Upvotes & 175.24 & 589.35 & 50.23 & N/A\\
        \hline
    \end{tabular}}}
    \caption{Statistics of our jokes dataset scraped from Reddit. Post length is measured as the number of tokens according to the Penn Treebank tokenizer.}
    \label{tab:data-stats}
\end{table}

\begin{table*}[ht!]
\setlength\tabcolsep{5pt}
\centering
{\renewcommand{\arraystretch}{1.2}
{\begin{tabular}{p{6.7cm}|p{3.5cm}|l|l}
Joke Title Text & Joke Body Text & Upvotes & Subreddit \\
\hline
Are you sweating while putting gas in your car? Feeling sick when paying for it? & You’ve got the Carownervirus & 227 & r/cleanjokes \\
Why was Stephen’s last name Hawking? & It’s not like he could be walking or talking & 115 & r/darkjokes \\
Did you hear about the mathematician that had sex? & He got sum! & 39 & r/DirtyJokes \\
\end{tabular}}}

\caption{Examples of Reddit jokes contained in our dataset. Posts contain title and body text, which concatenate to form a single coherent joke. Upvotes on a post indicate the popularity and humor value of that post within the subreddit's community. The jokes chosen for display are relatively less inappropriate. For a variety of other jokes from their original source, please see \href{https://www.reddit.com/r/cleanjokes/}{r/cleanjokes}, \href{https://www.reddit.com/r/darkjokes/}{r/darkjokes}, and \href{https://www.reddit.com/r/DirtyJokes/}{r/DirtyJokes}. We do not endorse these jokes.}
\label{table:joke-examples}
\end{table*}

\begin{figure}
    \centering
    \hspace*{-1.05em}
    \includegraphics[width=0.47\textwidth]{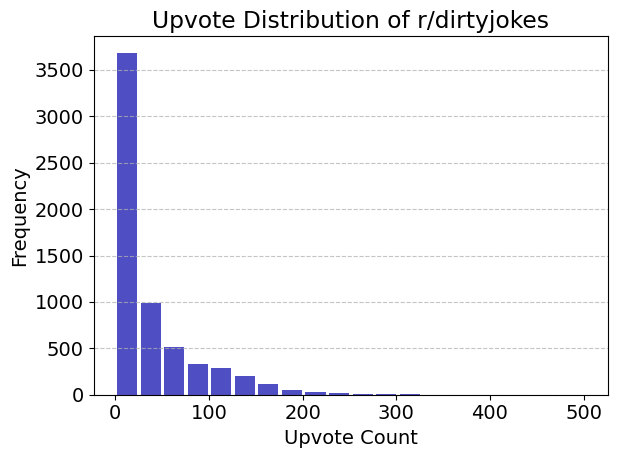}
    \caption{Long-tailed distribution of upvotes in r/dirtyjokes. The majority of jokes receive attain modest popularity, with a smaller subset of stand-out jokes achieve virality within the community. The other subreddits exhibit a similar distribution}
    \label{fig:long-tail}
\end{figure}

\section{A Humor Subtype Dataset}
To train the Naughtyformer, we introduce a dataset of 92,153 total jokes across categories of 1) Clean Jokes, 2) Dark Jokes, and 3) Dirty Jokes. We also include a fourth category, News, representing a non-joke. Table \ref{tab:data-stats} lists the statistics of our final dataset after collection and data processing. 

In particular, Clean jokes are jokes that are inoffensive in nature. Meanwhile, Dark Jokes are dark, morbid, cruel, offensive to some, or graphic in nature. Finally, Dirty Jokes are indecent jokes that consist of vulgar, sexist, racist, or discriminatory content. Critically, Dark and Dirty Jokes are distinct subtypes of humor.

We source all three joke types from Reddit, alongside news articles that act as non-jokes. A unique characteristic of Reddit posts is that post text decomposes into title text and body text. The title of a post oftentimes serves as a setup for the joke, and the body of the text is the punchline. Each post also consists of a certain number of \textit{upvotes}, or user reactions. Critically, this provides researchers with a human-centric metric for measuring the comical value of a joke in a given community. Figure \ref{fig:long-tail} displays an example distribution of upvote counts in a subreddits. Table \ref{table:joke-examples} outlines examples of jokes from each of our three subreddits. To represent a non-joke category, we use news articles from Thomson Reuters due to its reputation as a neutral, inoffensive media outlet.


\begin{table*}[ht!]
    \centering
    {\renewcommand{\arraystretch}{1.2}
    \begin{tabular}{ccccc}
        \hline
        \vspace{0.6em}
        \textbf{Model} & \textbf{Accuracy} & \textbf{Precision} & \textbf{Recall} & \textbf{F1 (Micro)} \\
        BERT-base & 86.33\% & 83.91\% & 83.26\% & 83.38\% \\
        RoBERTa-base & 86.70\% & 84.24\% & 84.13\% & 84.17\% \\
        DeBERTa-base (Naughtyformer) & \textbf{87.69\%} & \textbf{85.32\%} & \textbf{85.20\%} & \textbf{85.22\%} \\
        Longformer-base-4096 & 82.58\% & 80.94\% & 78.55\% & 78.64\% \\
        \hline
    \end{tabular}}
    \caption{Results of our finetuned BERT-base, RoBERTA-base, DeBERTa-base, and Longformer-base-4096 models on the 4-way humor subtype classification task.}
    \label{tab:results-2}
\end{table*}

\begin{table*}[htbp]
    \centering
    {\renewcommand{\arraystretch}{1.2}
    \begin{tabular}{ccccc}
        \hline
        \vspace{0.6em}
        \textbf{Model} & \textbf{Accuracy} & \textbf{Precision} & \textbf{Recall} & \textbf{F1 (Micro)} \\
        RoBERTa-base (TweetEval) & 75.84\% & 79.95\% & 71.05\% & 71.68\% \\
        DeBERTa-base (Naughtyformer) & \textbf{92.88\%} & \textbf{93.35\%} & \textbf{91.86\%} & \textbf{92.47\%} \\
        \hline
    \end{tabular}}
    \caption{Results on the offensive language detection task. We compare our most performant model finetuned on the humor subtype detection task, the DeBERTa-base model, with the offensive language detection model from the TweetEval benchmark \cite{barbieri2020tweeteval}.}
    \label{tab:results-3}
\end{table*}

\subsection{Subreddits as Natural Data}
Reddit is a social news website featuring user-curated feeds within well-defined communities aggregating specific content. We choose Reddit as our source of jokes precisely for these communities, also known as \textit{subreddits}. Subreddit users influence the popularity of a post by contributing upvotes. Notably, these subreddits feature explicit forum rules that gatekeep the type and content of posts that are allowed to appear in the forum. Moderators carefully determine if posts abide by the forum rules and fit the ethos of their given subreddit. Due to their siloed nature and community-specific content, subreddits thus act as a natural manifestation of clearly separated content categories. In particular, the we draw Clean, Dark, and Dirty Jokes from the subreddits r/cleanjokes, r/darkjokes, and r/DirtyJokes respectively. 

\subsection{Scraping Reddit}
We scrape the above three subreddits to obtain our joke dataset. Unfortunately, the official Reddit API limits user access to only the 1000 most recent listings in a given subreddit. To circumvent this, we use the Pushshift.io API as an effective surrogate. By sending paginated requests, sleeping between queries, continuously saving results, and multiprocessing queries, we successfully obtain all historical data from the three target subreddits.

\subsection{Data Processing}
After obtaining the scraped Reddit posts, we prune our dataset by removing posts that have a deleted or empty body of text, as well as duplicate posts. We accomplish this via RegEx to ensure robust removal. Additionally, because each joke post from the scraped subreddits consists of a separate title text and body text, we concatenate the title and body to form a full joke as a single string.






\section{Methods}
\subsection{Experiments}
We train and evaluate a set of large language models (see \ref{sec:models} for a full list) on our jokes dataset in order to classify each joke into one of 4 categories: Clean Joke, Dirty Joke, Dark Joke, and Not-a-Joke (news). We refer to this as the 4-way humor subtype classification task.

As a natural extension of our models and dataset, we also formulate an offensive language detection task. Given a body of text, we evaluate our models' ability to determine if the text contains offensive content. Since the Dark and Dirty Jokes in our dataset contain insensitive and inappropriate topics, we designate them to be offensive texts. Similarly, we consider instances of Clean Jokes and Not-a-Joke/News to be inoffensive texts. We refer to this binary classification problem as the offensive language detection task. 

\subsection{Metric}
To measure model performance, we calculate accuracy, precision, recall, and the micro-averaged F1 score. That is, we sum up the individual true positives, false positives, and false negatives of our system for different sets and compute the F1 score. We choose this metric in order to best reflect model performance on our uneven jokes distribution, which exhibits class imbalance.

\subsection{Models}
\label{sec:models}
We finetune a variety of pretrained large language models on our joke dataset. Specifically, we finetune BERT (110M params), RoBERTa (125M params), and DeBERTa (184M params). BERT \cite{devlin_bert_2019} is a multi-layer bidirectional transformer that can be finetuned with another additional output layer for a variety of downstream tasks. RoBERTa \cite{liu_roberta_2019} improves upon BERT, using the same architecture, but pretraining on an order of magnitude more data. DeBERTa \cite{He2020DeBERTaDB} further improves upon RoBERTa by using a disentangled attention mechanism and more parameters. Note the above models only support context lengths of up to 512 tokens. Texts greater than 512 tokens in length are truncated before being inputted into these architectures. 

However, 8.96\% of our dataset contains examples longer than 512 tokens, so we also evaluate the performance of Longformer-base-4096 (102M params) on our dataset. The Longformer \cite{beltagy_longformer_2020} employs an attention mechanism that scales linearly with sequence length, thus making it easy to process documents of much longer token length without unearable computational complexity. Critically, the Longformer is trained on and supports longer input sequences of up to 4,096 tokens, which makes it suitable for handling the longer texts in our dataset.



\subsection{Training}
We finetune our models using the AdamW optimizer \cite{loshchilov2018decoupled_adamw} with no weight decay. During training, we oversample minority classes to give every answer equal exposure. We reserve 20\% of our dataset as the test split. The remaining data are split 80\%-20\% to form the train and dev splits. We choose the learning rate and number of updates by grid search, using the dev split for validation. To increase experimentation speed, we enforce an early stopping callback.

\section{Results}
\subsection{Humor Subtype Detection}
The finetuned DeBERTa-base model -- dubbed the \textit{Naughtyformer} -- performs the best on humor subtype classification, achieving the highest accuracy, precision, recall and macro-F1 scores out of all models. Surprisingly, despite the Longformer's capability of accommodating larger text sequences, it performs the worst out of our four architectures. BERT, RoBERTa, and DeBERTa all perform better, despite only being able to leverage input text truncated to 512 tokens. We reason that this phenomena occurs because the majority of texts in our dataset are short and sit comfortably within the 512-token limit. Thus, the additional context seen by the Longformer is largely unhelpful and potentially distracting for classification purposes. Because we are the first to consider the humor subtype classification task, we are unable to compare our performance to prior work. Nonetheless, our full experimental results are shown in Table \ref{tab:results-2}.

\begin{figure}
    \centering
    \includegraphics[width=0.5\textwidth]{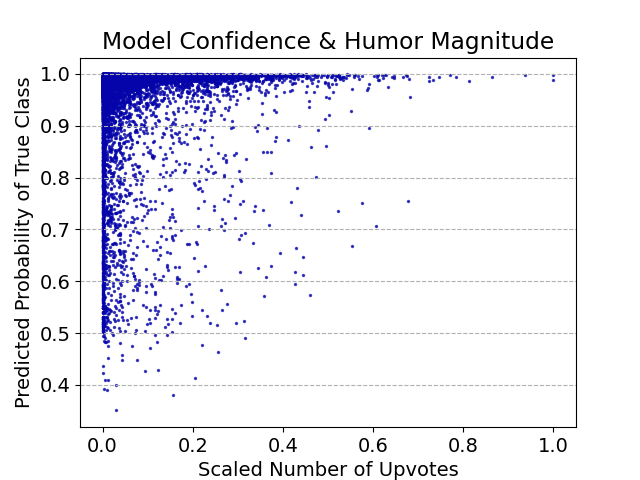}
    \caption{Our finetuned DeBERTa model's confidence in predicting the true humor subtype compared to the scaled number of post upvotes. All jokes in our dataset are plotted. More popular, and thus funnier, posts elicit more confident model responses in general.}
    \label{fig:model-confidence}
\end{figure}


\subsection{Offensive Language Detection in Humor}
We also evaluate the Naughtyformer on the offensive language detection task. Our finetuned DeBERTa model significantly outperforms the current state-of-the-art offensive language detection model from TweetEval \cite{barbieri2020tweeteval} by a $17.04\%$ increase in accuracy and $20.79\%$ increase in micro-F1. Full results are shown in Table \ref{tab:results-3}.

\subsection{Model Confidence \& Humor Magnitude}
To gauge our learned models' ability to measure a joke's magnitude of humor, i.e. \textit{how funny} it is, we investigate the relationship between model confidence and upvote count on a given post. First, we min-max scale the post upvotes according to the corresponding subreddit's upvote distribution. We then compare scaled upvotes to the Naughtyformer's probability, or confidence, of predicting the ground-truth humor subtype. Figure \ref{fig:model-confidence} displays the relationship between model confidence and humor magnitude. The Naughtyformer is sometimes highly confident on low upvote posts, though this is expected due to the heavily-skewed nature of the upvote distribution as exhibited in Figure \ref{fig:long-tail}. Critically, the more upvotes a post has, and equivalently the funnier it is, the more confident the model is in predicting the ground-truth humor class on average.  Based on these results, it appears the Naughtyformer can discern not only \textit{which} humor subtype a joke belongs to, but also \textit{how} funny it is.



\section{Conclusion}
We introduced a comprehensive dataset and model for classifying humor subtypes of the offensive variety. Furthermore, we investigated a novel perspective of offensive language detection and demonstrate that the Naughtyformer can detect offensive language in the context of jokes significantly better than state-of-the-art models. Finally, we showed that the Naughtyformer calibrates its classification confidence in alignment with human-centric measures of humor magnitude. Ultimately, we hope that our data and models can open up further research at the intersection of Natural Language Processing and Computational Social Science, and that our models can be used to mitigate overly offensive humor in the appropriate settings.

\section*{Limitations}



While the data and models in our work enable research on detecting humor subtypes and offensive language in jokes, it is important to note that the jokes appearing in our dataset may be found to be humorous (or more humorous) only by certain subsets of the population, such as members of the corresponding subreddits, and not necessarily by the general population.

Additionally, though our models are capable of discerning joke subtype and the presence of offensive language with high accuracy, here we do not investigate precisely what \textit{properties} cause a joke to belong to a certain class, or to be offensive. On this front, we encourage further research from the NLP \& CSS community to tackle the challenge of interpreting humor subtype and offensive humor detection.



\section*{Ethics Statement}

Our work presents a novel angle on the offensive language detection problem, specifically in the unexplored setting of offensive language contextualized within jokes. Advances in offensive language detection can lead to less hostile online environments, and we believe the onus is on us as researchers to contribute to this effort. That being said, we recognize that curating a dataset of offensive jokes may have adverse downstream effects if usage is not properly monitored. Nonetheless, in an age of ever-increasing online toxicity, hate speech, and polarization, we hope that our work can be leveraged to ensure a more harmonious online future. 


\bibliography{custom}




\end{document}